%% file: main.tex
\pdfoutput=1

\documentclass[11pt]{article}
\usepackage{authblk}
\usepackage{ACL2023}

\usepackage{times}
\usepackage{latexsym}

\usepackage[T1]{fontenc}

\usepackage[utf8]{inputenc}

\usepackage{microtype}

\usepackage{inconsolata}
\usepackage{microtype}
\usepackage{graphicx}
\usepackage{xcolor}
\usepackage{todonotes}
\usepackage{multirow}
\usepackage{amsmath,amssymb}
\usepackage{bbm}
\usepackage{array}
\usepackage[capitalise]{cleveref}
\usepackage{graphicx}
\usepackage{float} 
\usepackage{subfigure}
\usepackage{epstopdf}
\usepackage{tabularx}
\usepackage{algorithm}
\usepackage{algorithmic}
\usepackage{colortbl}
\usepackage{amsmath}
\usepackage{enumitem}
\usepackage{booktabs}
\usepackage{graphicx}
\usepackage{caption}

\title{Unsupervised Multi-document Summarization with Holistic Inference}

\author{\textbf{Haopeng Zhang}$^{1}$ \qquad \textbf{Sangwoo Cho}$^{2}$ \qquad \textbf{Kaiqiang Song}$^{2}$ \\ \bf
Xiaoyang Wang$^{2}$ \qquad Hongwei Wang$^{2}$ \qquad Jiawei Zhang$^{1}$ \qquad Dong Yu$^{2}$\\
\texttt{\{haopeng,jiawei\}@ifmlab.org}\\ \texttt{\{swcho,riversong,shawnxywang,hongweiw,dyu\}@global.tencent.com}\\
$^{1}$IFM lab, University of California, Davis \qquad $^{2}$ Tencent AI Lab, Bellevue, WA}

\begin{document}
\maketitle

\renewcommand{\thefootnote}{\fnsymbol{footnote}}
\footnotetext[1]{Work done during Haopeng Zhang's internship at Tencent AI Lab, Bellevue.}
\renewcommand{\thefootnote}{\arabic{footnote}}

\input{content/abstract.tex}
\input{content/introduction}
\input{content/related}

\input{content/method}

\input{content/experiment}

\input{content/analysis}
\input{content/conclusion.tex}
\section*{Limitations}
The proposed framework in this paper is mainly designed for low-resource scenarios without gold summaries for multi-document summarization. Adapting the framework for a supervised setting requires further investigation. Recently, large language models (LLM) like ChatGPT have shown strong zero-shot summarization ability, which may raise doubt about the necessity of unsupervised summarization methods. 

However, LLM suffers from the hallucination problem and MDS may exceed its input limit (e.g. 4,696 words for TAC) than the input limit of ChatGPT (500-word/4,000-character). In contrast, unsupervised summarization methods can tackle input of arbitrary length and have a faster inference speed than ChatGPT when processing long input documents. In addition, a recent study~\cite{zhang2023extractive} shows that ChatGPT's extractive summarization performance is still inferior to existing supervised systems in terms of ROUGE scores.

\section*{Ethical Consideration}

Our proposed framework forms summary by directly extracting sentences from source documents. Therefore, the extracted summary may be incoherent or contain unfactual co-references. In addition, the extracted summary will keep biased contents from the source sentences, if any.

\bibliography{anthology,custom}
\bibliographystyle{acl_natbib}

\end{document}

%% file: content/abstract.tex
\begin{abstract}
Multi-document summarization aims to obtain core information from a collection of documents written on the same topic. This paper proposes a new holistic framework for unsupervised multi-document extractive summarization. Our method incorporates the holistic beam search inference method associated with the holistic measurements, named Subset Representative Index (SRI). SRI balances the importance and diversity of a subset of sentences from the source documents and can be calculated in unsupervised and adaptive manners. To demonstrate the effectiveness of our method, we conduct extensive experiments on both small and large-scale multi-document summarization datasets under both unsupervised and adaptive settings. The proposed method outperforms strong baselines by a significant margin, as indicated by the resulting ROUGE scores and diversity measures. Our findings also suggest that diversity is essential for improving multi-document summary performance.

\end{abstract}

%% file: content/introduction.tex
\section{Introduction}
The multi-document summarization (\textbf{MDS}) is one of the essential tools to obtain core information from a collection of documents written for the same topic.
It seeks to find the main ideas from multiple sources with diversified messages. 
In spite of recent advances in MDS system designs \cite{mihalcea2004textrank,liu-lapata-2019-hierarchical,xiao-etal-2022-primera}, three major challenges hinder its development: 

First, existing extractive multi-document summarization systems rely on optimization with \textit{individual} scoring.
It becomes sub-optimal when we need to extract multiple summary sentences~\cite{zhong2020extractive}.
A typical individual system scores each candidate summary with only measurements of the newly added sentences during inference.
In contrast, the holistic system simultaneously measures all summary sentences and the relations among them.
Despite recent efforts in holistic methods on a single document~\cite{an2022colo,zhong2020extractive}, how to extract sentences holistically for multi-document summarization remains open. 
In this work, we propose an inference method that holistically optimizes the extractive summary under multi-document setting.

\begin{figure}[!tbp]
\centering
\includegraphics[width=0.5\textwidth]{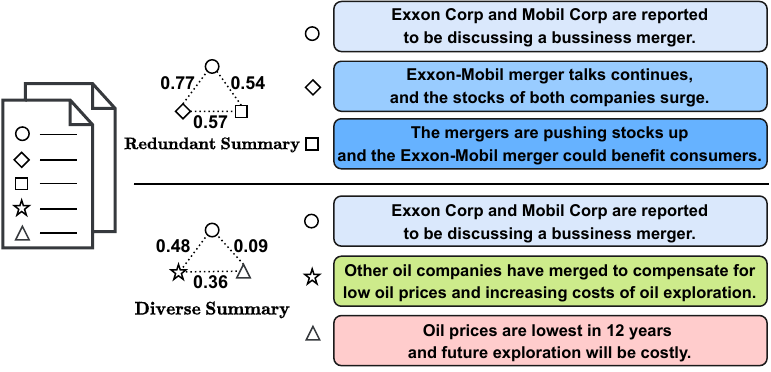}
\caption{An example of a diverse summary vs. a redundant summary. Sentences in the redundant summary have higher semantic similarity than a diverse summary.}
\label{fig:intro}
\end{figure}

Second, multi-document summarization naturally contains excessively redundant information~\cite{lebanoff2018adapting}.
An ideal summary should provide important information with diversified perspectives~\cite{Nenkova:2011}.
In Figure~\ref{fig:intro}, we show a salient and diversified summary versus a salient but redundant summary.
A salient and diversified summary often covers the information thoroughly, while a salient but redundant summary is usually incomplete. 
Different from existing approaches~\cite{suzuki-nagata-2017-cutting,cho-etal-2019-multi,xiao2020systematically} for limiting the repetitions, we introduce \textbf{S}ubset \textbf{R}epresentative \textbf{I}ndex (\textbf{SRI}), a holistically balanced measurement between importance and diversity for extractive multi-document summarization.

Finally, recent deep learning-based supervised summarization methods are data-driven and require a massive number of high-quality summaries in the training data.
Nevertheless, hiring humans to write summaries is always expensive, time-consuming, and thus hard to scale up.
This problem becomes more severe for multi-document summarization, since it requires more effort to read more documents.
Therefore, existing multi-document summarization datasets are either small-scale \cite{Over:2004,Dang:2008} or created by acquiring data from the Internet with automatic alignments~\cite{fabbri-etal-2019-multi,antognini-faltings:2020:LREC2} that could be erroneous.
Here we propose an unsupervised multi-document summarization method to tackle the low-resource issue.
It can further benefit the unsupervised multi-document summarization, with the adaptive setting using large-scale high-quality single-document summarization data (e.g., CNN/DailyMail~\cite{hermann2015teaching}).

In this work, we present a novel framework for unsupervised extractive multi-document summarization, aiming to holistically select the extractive summary sentences. 
The framework contains the holistic beam search inference method associated with the holistic measurements named \textbf{SRI} ( \textbf{S}ubset \textbf{R}epresentative \textbf{I}ndex).
The SRI is designed as a holistic measurement for balancing the importance of individual sentences and the diversity among sentences within a set.
To address data sparsity, we propose to calculate SRI in both unsupervised and adaptive manners.
Unsupervised SRI relies on the centrality from graph-based methods ~\cite{Erkan:2004,mihalcea2004textrank} for subset importance measurement, while adaptive SRI uses BERT \cite{devlin2018bert} fine-tuned on single document summarization (SDS) corpus for sentence importance measurement.
Our method shows performance improvements in both the summary informativeness and diversity scores, indicating our approach can achieve better coverage of documents while maintaining the gist information of multi-documents. We highlight the contributions of our work as follows:
\begin{itemize}
    \item We propose a novel holistic framework for multi-document extractive summarization. Our framework incorporates a holistic inference method for summary sentence extraction and a holistic measurement called Subset Representative Index (SRI) for balancing the importance and diversity of a subset of sentences.

    \item We propose two unsupervised ways to measure SRI by using graph-based centrality or adapting from a single document corpus.
    \item 
We conduct extensive experiments on several benchmark datasets, and the results demonstrate the effectiveness of our paradigm under both unsupervised and adaptive settings. Our findings suggest that effectively modeling sentence importance and pairwise sentence similarity is crucial for extracting diverse summaries and improving summarization performance. 
\end{itemize}

%% file: content/related.tex
\section{Related Works}

\paragraph{Multi-document Summarization} 
Traditional non-neural approaches to multi-document summarization have been both extractive~\cite{carbonell1998use,Erkan:2004,mihalcea2004textrank} and abstractive \cite{Ganesan:2010}. Recent neural MDS systems rely on Transformer-based encoder-decoder model to process the integrated long documents with hierarchical inter-paragraph attention~\cite{liu-lapata-2019-hierarchical,fabbri-etal-2019-multi}, or attention across representations of different granularity~\cite{jin2020multi}. This work focuses on unsupervised MDS scenarios where gold reference summaries are unavailable. Prior unsupervised MDS systems are mostly graph-based~\cite{Erkan:2004,liu2021unsupervised}. Similar to our adaptive setting, \citet{lebanoff2018adapting} proposed to adapt the encoder-decoder framework from a single document corpus, but our work focuses on extractive summarization setting with holistic inference.

\begin{figure}[!tbp]
\centering
\includegraphics[width=0.49\textwidth]{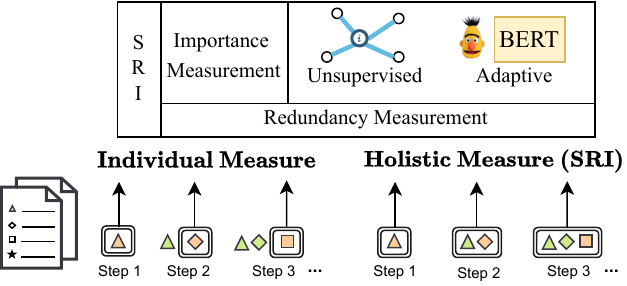}
\caption{Illustration of the proposed holistic framework for multi-document summarization. The individual inference only resorts to each candidate while the holistic inference is based on all candidates. Orange and Green indicate newly added sentences and already added ones to the summary respectively.}
\label{fig:pipeline}
\end{figure}

\paragraph{Sentence Importance Measurements} 
Most works formulate extractive summarization as a sequence classification problem and use sequential neural models with different encoders like recurrent neural networks~\cite{cheng2016neural,nallapati2016abstractive} and pre-trained language models~\cite{liu2019text,zhang2023diffusum}. The prediction probabilities are treated as the importance measurement of sentences. On the other hand, unsupervised graph-based methods calculate the importance of sentences with node centrality and rank them for the summaries, including TextRank~\cite{mihalcea2004textrank}, LexRank~\cite{Erkan:2004}, \textsc{PacSum}~\cite{zheng2019sentence}, and its variants \cite{liang2021improving,liu2021unsupervised}.
Recent researches~\cite{ xu2019discourse,wang-etal-2020-heterogeneous,zhang-etal-2022-hegel,zhang-etal-2023-contrastive-hierarchical} have explored Graph Neural Networks to obtain better representations for each sentence.
Graph methods have merits in considering implicit document structure and to adapt with regardless of the input length. 

graph neural network message passing or simply

\paragraph{Redundancy} Considering only the importance of sentences for the summary leads to repeated information, and resolving the redundant contents is an essential problem in the extractive summarization system.
Traditional methods to tackle redundancy relies on discrete optimization problem like Maximal Marginal Relevance (MMR)~\cite{carbonell1998use}, Determinantal Point Process (DPP)~\cite{Kulesza:2012}, and submodular selection~\cite{lin2009graph}. 
Trigram blocking is introduced to explicitly reduce redundancy by avoiding sentences that share a 3-gram with the previously added one~\cite{liu2019text}. 
\citet{paulus2017deep} first adopt trigram blocking in decoding for abstractive summarization. \citet{ma2016unsupervised} proposed the sentence filtering and beam search methods for extractive summarization sentence selection.
\citet{xiao2020systematically} conducted a systematic study of redundancy in long documents.

%% file: content/method.tex
\section{Method}
This section provides a detailed description of our proposed holistic MDS summarization framework.
We first explain how we formulate the MDS problem holistically in \Cref{sec:problem_formulation}.
The overall architecture of our holistic framework is shown in Figure~\ref{fig:pipeline}, which includes holistic inference methods for summary sentence extraction in~\Cref{sec:inference}, and a new holistic measurement, the Subset Representative Index (SRI) in~\Cref{sec:measurements}.
 
\subsection{Problem Formulation}
\label{sec:problem_formulation}
Multi-document summarization typically takes a collection of $n$ documents $\mathcal{D} = \{D^{(1)}, \dots, D^{(n)}\}$ as inputs.
Each document contains a varying number of sentences $D^{(i)} = \{s^{(i)}_{0}, \dots, s^{(i)}_{l_i}\}$, where $l_i$ is the number of sentences in the $i$-th document.
Let $\mathcal{S}$ be the collection of all sentences, i.e. $\mathcal{S}=D^{(1)} \cup \dots \cup D^{(n)}$.
Additionally, let $e_{i,j}$ denote the similarity score between sentence $s_i$ and sentence $s_j$.
Our goal is to select a representative subset of sentences $\mathcal{S'} \subset \mathcal{S}$ that maximizes the total importance of the subset while minimizing the redundancy within sentences in the subset at the same time.

\subsection{Holistic Inference}
\label{sec:inference}

Most existing approaches for unsupervised extractive summarization formulate it as an individual sentence ranking problem.
They first calculate a measurement $\mathcal{M}(s_i)$ (e.g. sentence importance) for each sentence $s_i \in \mathcal S$ and rank all sentences in $\mathcal S$ accordingly.
For summary inference, they directly use an \textit{individual greedy} method that adds one sentence with the highest ranking at a time until the desired total number of summary sentences is reached.
In contrast, a holistic summarization method should evaluate a subset of sentences $\mathcal{M}(\mathcal{S'})$ as a whole, then select the best subset $\mathcal{S'}$.
The setting formulates the holistic summary inference into a best subset selection problem, which has exponential time complexity.

To address the exponential time complexity issue, we propose several holistic inference methods for summary sentence extraction.
These methods optimize subsets of sentences using subset measurements, as opposed to the individual greedy inference method.
We describe the different variants of the proposed method as below.

\paragraph{Holistic Greedy Method.}
The most straightforward way to address the exponential time complexity issue is to adopt a greedy approach.
Similar to the individual greedy method, the holistic greedy method also adds one sentence at a time. However, it picks the sentence using a subset measurement that takes into account the previously selected sentences.
Formally, at each step, the method selects the sentence that maximizes the following objective:
\begin{equation}
\operatorname{argmax}_{s_i \in \mathcal{S} \setminus \mathcal{S'}} \mathcal{M}(\mathcal{S'} \cup \{s_i\}),
\end{equation}
where $\mathcal{S'}$ represents the previously selected sentences.

\paragraph{Holistic Exhaustive Search.}
It is a brute-force method that considers every possible subset with the desired number of sentences.
However, due to the exponential computation time, it is necessary to first filter out low-importance candidates using $\mathcal{M}(\{s_i\})$ to reduce the search space.

\paragraph{Holistic Beam Inference}.
We also propose Holistic Beam Inference which balances the trade-off between search space size and efficiency.
It is a more advanced holistic inference method that adapts the beam-search decoding algorithm.
We illustrate the algorithm in \Cref{alg:beam}.
At each step, it considers the top-k candidate subsets, which enlarges the search space and therefore has a higher chance of finding a better subset solution compared to the holistic greedy method.
Meanwhile, the algorithm has linear time complexity, making it more efficient than the holistic exhaustive search method.

\subsection{Subset Representative Index}
\label{sec:measurements}
To complement the holistic inference methods, we propose a new subset measurement, Subset Representative Index (SRI), denoted as $\mathcal{M}(\mathcal{S'})$.
It balances the importance measurement $\mathcal{I}(\mathcal{S'})$ and redundancy measurement $\mathcal{R}(\mathcal{S'})$.

An ideal extractive summary should select the most representative subset from a collection of the input sentences, maximizing the total non-redundant salient information passed to the user.
SRI is a holistic subset measurement that balances the importance and redundancy of a subset of sentences from the source documents.
Formally, we define SRI as below:
\begin{align}
    \mathcal{M}(\mathcal{S'}) =& \mathcal{I}(\mathcal{S'}) - \lambda \cdot \mathcal{R}(\mathcal{S'}),
\end{align}
where $\mathcal{I}(\mathcal{S'})$ measures the informativeness of a set of sentences, and $\mathcal{R}(\mathcal{S'})$ measures the redundancy within the set. The parameter $\lambda$ is used to control the weight of the redundancy in the overall SRI score. We detail the methods for measuring the set importance and redundancy in an unsupervised manner as follows.

\input{tables/alg_beam}
\paragraph{Graph-Based Importance Measurement.~}
To measure the importance of sentences, we use a graph-based approach.
We construct a graph $\mathcal{G}=( \mathcal{V}, \mathcal{E} )$, where node $v_i \in \mathcal{V}$ represents sentence $s_i \in \mathcal{S}$, and edge $e_{i,j} \in \mathcal{E}$ represents the similarity between sentence $s_i$ and $s_j$.
Our proposed approach for sentence similarity score employs a combination of two methods: TF-IDF and Sentence-BERT \cite{reimers-2019-sentence-bert}.
TF-IDF is used to encode sentences with surface-form similarity, while Sentence-BERT is used to encode sentences with semantic similarity:
\begin{equation}
\label{equ:refernece_of_equ_1}
e_{i,j} = \alpha \cdot \mathbf{c}_i^\top \mathbf{c}_{j} + (1-\alpha) \cdot \mathbf{r}_i^\top \mathbf{r}_{j},
\end{equation}
where $\mathbf{c}_i$, $\mathbf{c}_j$, $\mathbf{r}_i$ and $\mathbf{r}_j$ are the corresponding TF-IDF features and sentence embeddings for the $i$-th and $j$-th sentences, respectively.
The weight term $\alpha \in [0, 1]$ is a configurable hyperparameter to balance between statistical similarity and contextualized similarity.

Inspired from \cite{ mihalcea2004textrank, Erkan:2004}, we define the importance of a sentence as its node centrality in the graph, which is calculated as the sum of the weights of edges connected to the node representing this sentence:
\begin{align}
    \label{equ:refernece_of_equ_2}
    \mathcal{I}(s_i)= \sum_{s_j \in \mathcal{S} \backslash s_i}{e_{i,j}}.
\end{align}
Similarly, the importance of a subset of sentences is defined as the total weights between the subgraph and the remaining graph:
\begin{align}
    \mathcal{I}(\mathcal{S'}) =& \frac{1}{|\mathcal{S}| - |\mathcal{S'}|}\sum_{s_i \in \mathcal{S'}, s_j \in \mathcal{S} \setminus \mathcal{S'}}{e_{i,j}}  \nonumber\\ \approx&\frac{1}{|\mathcal{S}|}\sum_{s_i \in \mathcal{S'}, s_j \in \mathcal{S} \setminus \mathcal{S'}}{e_{i,j}}.
\end{align}
Since $|\mathcal{S'}|$ is usually far smaller than $|\mathcal{S}|$ in summarization tasks, we can approximate the denominator by using $|\mathcal{S}|$ directly. This way, the subset importance only takes into account the relationship of the subset with the remaining sentences, rather than considering dependencies within the subset.

\paragraph{Adaptive Importance Measurement.~} 
In spite of the data sparsity issue in MDS, the Single Document Summarization (SDS) task has abundant high-quality labeled data \cite{hermann2015teaching,narayan2018don,cohan2018discourse}.
We propose a method called adaptive importance measurement, which adapts SDS data for MDS importance measurement.
This method utilizes the labeled data from SDS to train a model for predicting the importance of sentences in MDS.

In the adaptive setting, we fine-tune the BERT \cite{devlin2018bert} to a sentence importance scorer on SDS datasets and then adapt the fine-tuned model to the target MDS datasets. Specifically, we first calculate the normalized salience of a sentence as:
\begin{equation}
\begin{gathered}
f\left(s_i\right)=\mathbf{v}^\top \tanh \left(\mathbf{W}_1 \mathbf{r}_i\right), \\
\text {salience}\left(s_i\right)=\frac{f\left(s_i\right)}{\sum_{s_j \in D} f\left(s_j\right)}, \\
\end{gathered}
\end{equation}
where $\mathbf{W}$ is a trainable weight, and $\mathbf{r}_i$ is the contextualized representation of sentence $s_i$.
Then, we fine-tune BERT to minimize the following loss:
\begin{equation}
\begin{gathered}
R\left(s_i\right)=\operatorname{softmax}\left( \text{ROUGE}\left(s_i\right)\right),\\
\mathcal{L}=-\sum_D \sum_{s_i \in D} R\left(s_i\right) \log \text {salience}(s_i). 
\end{gathered}
\end{equation}
The fine-tuned BERT can be directly adapted to the MDS datasets and calculate the adaptive importance measurement for sentences.

\paragraph{Redundancy Measurement.~}
The redundancy measurement for a subset of sentences $\mathcal{S'}$ is defined as the total similarity score of each sentence with its most similar counterpart. This measurement captures the degree of overlap between the sentences in the subset, indicating the level of redundancy present in the selected sentences:

\begin{equation}
\mathcal{R}(\mathcal{S'})=\sum_{s_i \in \mathcal{S'}} {\max_{s_j \in \mathcal{S'} \setminus \{s_i\}}{e_{i,j}}}.
\end{equation}

Overall, we can calculate SRI in both unsupervised and adaptive manners.
Our holistic framework extracts summaries as a whole with the holistic inference method, which is guided by SRI to measure the importance and redundancy of a subset of sentences.
This approach allows us to balance the importance and redundancy of a summary, making it more informative and coherent.

%% file: tables/alg_beam.tex
\begin{algorithm}[tb]
\small
\caption{Holistic Beam Inference}
\label{alg:beam}
\textbf{Input}: set of sentences $\mathcal{S}$, Measurement $\mathcal{M}(\cdot)$  \\
\textbf{Parameter}: \# summary sentences $N<|\mathcal{S}|$, beam size $k$\\
\textbf{Output}: the selected subset $\mathcal{S'}$
\begin{algorithmic}[1] 
\STATE The candidate set $\mathcal{C} \gets \{ \varnothing$\} 
\FOR{$N$ times}
    \STATE The beam set $\mathcal{C'} \gets \{ \varnothing$\} 
    \FOR{$\mathcal{X} \in \mathcal{C}$ } 
        \STATE $\mathcal{X'} \gets \text{arg-top-}k\text{-max}_{s \in \mathcal{S} \setminus \mathcal{X}} {\mathcal{M}(\mathcal{X} \cup \{s\})}$
        \FOR {$x \in \mathcal{X'}$}
        \STATE Add $\mathcal X \cup \{x\}$ to $\mathcal C'$
\ENDFOR
    \ENDFOR
    \STATE $\mathcal{C} \gets \text{arg-top-}k\text{-max}_{\mathcal X \in \mathcal{C'}}{\mathcal{M}(\mathcal X)}$
\ENDFOR
\STATE \textbf{return} $\operatorname{argmax}_{\mathcal{X} \in \mathcal{C}}{\mathcal{M}(\mathcal{X})}$
\end{algorithmic}
\end{algorithm}

%% file: content/experiment.tex
\section{Experiments}
\label{sec:experiments}

In this section, we provide details on our experimental setup, including the datasets, evaluation metrics, baselines, and implementation details (Section~\ref{sec:setting}). We then present the results of our model on benchmark MDS datasets in both unsupervised (Section~\ref{sec:un_summ_results}) and adaptive (Section~\ref{sec:zero_summ_results}) settings.

\subsection{Experimental Setting}
\label{sec:setting}

\noindent \textbf{Dataset.~}
We evaluate our unsupervised method on benchmark multi-document summarization datasets.
Particularly, we use MultiNews~\cite{fabbri-etal-2019-multi}, WikiSum~\cite{Liu2018GeneratingWB}, DUC-04~\cite{Over:2004}, and TAC-11~\cite{Dang:2008} datasets.
MultiNews is collected from a diverse set of news articles on newser.com.
It is a large-scale dataset containing reference summaries written by professional editors.
WikiSum is another large-scale dataset that provides documents and summaries from Wikipedia webpages where the documents come from the reference webpages of Wikipedia articles and top-10 Google searches, and the summaries are the lead section of the Wikipedia articles.
We use the top-40 high-ranked paragraphs for the document inputs following~\cite{liu-lapata-2019-hierarchical}.

For summary extraction, we use the average number of reference sentences: 10 and 5, respectively on MultiNews and WikiSum. 
For the DUC and TAC datasets, the task is to generate a succinct summary of up to 100 words from a set of 10 news articles.
We report results on DUC-04 and TAC-11, which are standard test sets used in previous studies~\cite{hong-etal-2014-repository,cho-etal-2019-improving}.
DUC-03 and TAC-08/09/10 are used for the validation set to tune hyper-parameters. For adaptive setting, we fine-tune BERT on single document summarization dataset CNN/DailyMail~\cite{hermann2015teaching} and directly adapt to MDS test sets.
Table~\ref{datasets} shows the statistics of the datasets in detail.

\input{tables/dataset}
\input{tables/result_unsup.tex}
\noindent \textbf{Evaluation Metrics.~}
The extracted summaries are evaluated against human reference summaries using ROUGE~\cite{lin2004rouge}\footnote{w/ options 
\textsf{-n 2 -m -w 1.2 -c 95 -r 1000 -l 100} for DUC/TAC} for the summarization quality.
We report ROUGE-1, ROUGE-2, ROUGE-SU4, and ROUGE-L\footnote{Due to some legacy issues, some baselines report the original ROUGE-L, others report ROUGE-Lsum.} that respectively measure the overlap of unigrams, bigrams, skip bigrams with a maximum distance of 4 words, and the longest common sequence between extracted summary and reference summary.
To align with previous works, we report R-1, R-2, R-L for Multinews and Wikisum datasets, and R-1, R-2, R-SU4 for DUC and TAC datasets. 
For all baseline methods, we report ROUGE results from their original papers if available or use results reported in~\cite{cho-etal-2019-improving, liu2021unsupervised}.
We also report the measure of diversity for the generated summaries by calculating a unique n-gram ratio~\cite{xiao2020systematically,peyrard2017learning} defined as:
\begin{equation}
\text {uniq $n$-gram ratio}=\frac{\#\text { uniq-$n$-gram }}{ {\#}n\text {-gram }}
\end{equation}

\noindent \textbf{Baselines.~}
We compare our methods with strong unsupervised summarization baselines.
In particular, \textit{MMR}~\cite{carbonell1998use} combines query relevance with information novelty in the context of summarization. \textit{LexRank}~\cite{Erkan:2004} computes sentence importance based on eigenvector centrality in a graph representation of sentences.
\textit{TextRank}~\cite{mihalcea2004textrank} adopts PageRank~\cite{page1999pagerank} to compute node centrality recursively based on a Markov chain model. \textit{SumBasic}~\cite{vanderwende2007beyond} is an extractive approach assuming words frequently occurring in a document cluster are more likely to be included in the summary.
\textit{KL-Sum}~\cite{haghighi2009exploring} uses a greedy approach to add a sentence to the summary to minimize the KL divergence. \textit{PRIMERA}~\cite{xiao-etal-2022-primera} is a pyramid-based pre-trained model for MDS that achieves state-of-the-art performance. We compare it under its zero-shot setting.


%

\noindent \textbf{Implementation Details.~}
We run all experiments with 88 Intel(R) Xeon(R) CPUs.
We combine the surface indicator based on TF-IDF and contextualized embeddings. We treat each document cluster as a corpus and each sentence as a document when calculating the TF-IDF scores. We employ the pre-trained sentence-transformer~\cite{reimers-2019-sentence-bert} and extract sentence representations using a checkpoint of `all-mpnet-base-v2'.

The graph edges with low similarity are treated as disconnected to emphasize the connectivity of the graph and avoid noisy edge connections.
We keep a threshold $\tilde{e}$ for edge weights such that edges with similarity scores smaller than $\tilde{e}$ will be set to $0$.
Here $\tilde{e}$ is controlled by a hyper-parameter to be tuned according to datasets.
The final representation of edge weight between two sentences $(s_i,s_j)$ is
\begin{equation}
    {e}_{i,j}= \max(\text{sim}(s_i, s_j)- \tilde{e}, 0),
\end{equation}
where $\tilde{e} = \min({e})+\theta \;(\max({e})-\min({e}))$ is the threshold controlled by hyper-parameter $\theta$.
For exhaustive search, we filter out the sentences with low centrality and only keep the top $15$ sentences at inference.

All hyper-parameters are tuned on validation sets on MultiNews and WikiSum and training sets on DUC and TAC.
The best parameters are selected based on the highest R-1 score.
More specific, for the balancing factor $\lambda$ in SRI, we use $\{2^{-13}$, $2^{-7}$, $2^{-4}$, $2^{-6}\}$ on DUC, TAC, MultiNews and WikiSum dataset.
For $\alpha$ that weighted the contributions of TF-IDF and contextualized sentence similarity, we use $0.9$ on News domain datasets and $0.8$ on the WikiSum dataset.
The edge weight threshold $\theta$ is $\{0$, $0$, $0.1$, $0.1\}$ for DUC, TAC, MultiNews and WikiSum.
As for beam search, we use beam size  $\{4$, $4$, $4$, $3\}$ on the corresponding datasets.

\subsection{Unsupervised Summarization Results}
\label{sec:un_summ_results}
The unsupervised summarization results on four benchmark MDS datasets are shown in Table~\ref{tab:results_unsup}.

The summarization performance of our method outperforms strong unsupervised baselines.
Note that MultiNews and WikiSum datasets provide abundant training samples and contain shorter input than the DUC or TAC datasets.
Our method performs better than the pre-trained model, \textit{PRIMERA} with a zero-shot setting. Compared to the baseline (Sent. greedy) that extracts sentences solely based on importance, balancing diversity with SRI boosts performance by a large margin.

For the DUC-04 and TAC-11 datasets, our proposed methods outperform unsupervised baselines by a large margin. It demonstrates that balancing the summary informativeness and diversity during the sentence extraction process is crucial for better summary quality.
Note that the input length of DUC/TAC datasets is extremely long spanning an average of $180$ sentences.
These long input easily exceeds the input capacity of transformer-based models possibly resulting in information loss from documents. The proposed methods on the other hand process documents regardless of the input length or formats (SDS or MDS). Also, our unsupervised methods have the advantage of processing datasets with small training data.
The supreme performances on datasets with different input lengths and low-resource data illustrate the effectiveness of our methods. To further verify the model performance, we also conduct a human evaluation by experts on a scale of $5$. The results shown in Table~\ref{tab:human} also prove our method outputs better summaries in an unsupervised setting.
\input{tables/human}
\input{tables/result_zero.tex}
\subsection{Adptive Summarization Results}
\label{sec:zero_summ_results}

The experimental results under the adaptive setting are shown in Table~\ref{tab:results_zero}. Compared to large pre-trained generation models (BART) and other task-specific pre-trained summarization models (PEGASUS, PRIMERA), our framework shows strong performance when adapting from a single document summarization dataset. We also notice fine-tuning on single document summarization corpus improves the performance of all pre-trained models, but still, our framework achieves the best results under the adaptive setting.

%% file: tables/dataset.tex
\begin{table}[!tbp]
\setlength{\tabcolsep}{2.2pt}
\renewcommand{\arraystretch}{1.25}
\centering
\begin{small}
\begin{tabular}{cccccccc}
\hline Dataset & \# test & \# ref. & avg.w/doc & avg.w/sum \\
\hline 
DUC-04 & $50$ & $4$& $4,636$  & $109.6$  \\
TAC-11 & $44$ & $4$& $4,696$ & $99.7$ \\
Multi-News & $ 5,622$ & $1$ & $2,104$  & $264.7$  \\
Wikisum & $38,205 $& $1$ & $2,800$  & $139.4$  \\
CNNDM(SDS) &11,489&1& 766.1&58.2\\

\hline
\end{tabular}
\end{small}
    \caption{Detailed statistics of four multi-document datasets. $\#$test denotes the number of document clusters in the test set, $\#$ref denotes the number of reference summaries, avg.word(doc) denotes the average number of words in the source document cluster, avg.word(sum) denotes the average number of words in the ground truth summary.}
    \label{datasets}
\end{table}

%% file: tables/result_unsup.tex
\begin{table*}[t]
\setlength{\tabcolsep}{5pt}
\renewcommand{\arraystretch}{1.25}
\centering
\begin{small}\scalebox{1}{
\begin{tabular}{l|ccc|ccc|ccc|ccc}
\hline
& \multicolumn{3}{c|}{\textbf{\textsl{DUC-04}}}& \multicolumn{3}{c|}{\textbf{\textsl{TAC-11}}}& \multicolumn{3}{c|}{\textbf{\textsl{MultiNews}}} & \multicolumn{3}{c}{\textbf{\textsl{WikiSum}}}\\
\textbf{System} & \textbf{R-1} & \textbf{R-2} & \textbf{R-SU}  & \textbf{R-1} & \textbf{R-2} & \textbf{R-SU}& \textbf{R-1} & \textbf{R-2} & \textbf{R-L} & \textbf{R-1} & \textbf{R-2} & \textbf{R-L*} \\
\hline

\rowcolor{gray!10}
\multicolumn{13}{c}{\textbf{\textsl{Unsupervised Systems}}} \\
\hline
\textsc{Lead} &30.77&8.27&7.35& 32.88&7.84&11.46&   39.41 & 11.77 & 14.51 & 37.63 & 14.75 & 33.76 \\
MMR~\citeyearpar{carbonell1998use}&30.14&4.55&8.16&31.43&6.14&11.16&             38.77 & 11.98 & 12.91 & 31.22 & 10.24 & 22.48 \\
LexRank~\citeyearpar{Erkan:2004} & 34.44 & 7.11 & 11.19  & 33.10 & 7.50 & 11.13&         38.27 & 12.70 & 13.20 & 36.12 & 11.67 & 22.52 \\
TextRank~\citeyearpar{mihalcea2004textrank}&33.16&6.13&10.16&33.24&7.62&11.27&        38.44 & 13.10 & 13.50 & 23.66 & 7.79 &  21.23 \\
SumBasic (\citeyear{vanderwende2007beyond})&   29.48 & 4.25 & 8.64  & 31.58 & 6.06 & 10.06 &-&-&-&-&-&-\\
KLSumm (\citeyear{haghighi2009exploring})&31.04&6.03&10.23&31.23&7.07&10.56&-&-&-&-&-&-\\
PRIMERA~\citeyearpar{xiao-etal-2022-primera}&35.10&7.20&17.90&-&-&-&         42.00 & 13.60 & 20.80 & 28.00 & 8.00 & 18.00 \\
Individual. Greedy&34.81&7.85&11.37&34.42&8.10&11.25&40.48&13.49&16.14&37.24&10.29&32.77\\
\hline
\rowcolor{gray!10}
\multicolumn{13}{c}{\textbf{\textsl{Our Methods}}} \\
\hline
SRI+beam &\textbf{36.84}&\textbf{8.37}&{12.28}&\textbf{35.37}&\textbf{8.49}&\textbf{11.73}& \textbf{44.22} & \textbf{14.63} & 18.61 & 38.94& 15.23 & 34.12 \\
SRI+exh &{36.70}&\textbf{8.37}&\textbf{12.31}&{35.19}&{8.31}&{11.34}& 43.16 &  {14.58} & {18.00} &\textbf{39.26} & \textbf{16.15} & \textbf{34.19} \\
\hline
\end{tabular}} 
\end{small}
\caption{ROUGE-F1 scores on four datasets under the unsupervised setting. Best unsupervised results are bold. For a fair comparison, we report {R-L} on Multinews and {R-Lsum}~\cite{see2017get} for WikiSum and limit summaries to 100 words on DUC-04 and TAC-11. R-L are marked with * if reporting ROUGE-
Lsum numbers.
}
\label{tab:results_unsup}
\end{table*}

%% file: tables/human.tex
\begin{table}[t]
\setlength{\tabcolsep}{2.2pt}
\renewcommand{\arraystretch}{1.25}
\centering
\begin{small}
\begin{tabular}{l|cccc}
\hline
\textbf{Method}  & \textbf{Fluent} & \textbf{Informative} & \textbf{Faithful} &\textbf{Overall}  \\
\hline
MMR & 3.2 &3.5&\textbf{4.7}  &3.2 \\
PRIMERA & \textbf{4.3} & {2.5}&3.3 & {3.3} \\ 
SRI & 3.8  & \textbf{4.3}&\textbf{4.7} & \textbf{4.0}\\

\hline
\end{tabular}
\end{small}
\caption{{Human evaluation} results on a scale of 1-5.
}
\label{tab:human}
\end{table}

%% file: tables/result_zero.tex
\begin{table}[t]
\setlength{\tabcolsep}{2.2pt}
\renewcommand{\arraystretch}{1.25}
\centering
\begin{small}\scalebox{1}{
\begin{tabular}{l|ccc|ccc}
\hline
& \multicolumn{3}{c|}{\textbf{\textsl{DUC-04}}} & \multicolumn{3}{c}{\textbf{\textsl{MultiNews}}}\\
\textbf{System} & \textbf{R-1} & \textbf{R-2} & \textbf{R-L}  & \textbf{R-1} & \textbf{R-2} & \textbf{R-L}  \\
\hline

\rowcolor{gray!10}
\multicolumn{7}{c}{\textbf{\textsl{Adaptive Systems}}} \\
\hline
BART(\citeyear{lewis2019bart}) &24.1
& 4.0&15.3&27.3&6.2&15.1\\
BART (CNNDM) &29.4
& 6.1&16.2&36.7&8.3&17.2\\
PEGASUS (\citeyear{zhang2020pegasus}) & 32.7&7.4&17.6&32.0&10.1&16.7\\
PEGASUS(CNNDM)  & 34.2&7.5&17.4&35.1&11.9&18.2\\
LED(\citeyear{Beltagy2020Longformer}) &16.6&3.0&12.0&17.3&3.7&10.4\\
PRIMERA (\citeyear{xiao-etal-2022-primera}) 
&35.1&7.2&17.9&42.0&13.6&20.8\\
\hline
\rowcolor{gray!10}
\multicolumn{7}{c}{\textbf{\textsl{Our Systems}}} \\
\hline
SRI+beam (graph) &{36.8}&{8.4}&{16.4}&{44.2} & \textbf{14.6} & 18.6 \\
SRI+beam (CNNDM) &\textbf{36.9}&\textbf{8.6}&\textbf{18.5}&\textbf{44.6}&{14.3}&\textbf{21.1}\\

\hline
\end{tabular}}
\end{small}
\caption{ROUGE-F1 results on DUC-04 and Multinews datasets under the adaptive setting. Models adapted from CNN/DailyMail dataset are marked in the bracket. 
}
\label{tab:results_zero}
\end{table}

%% file: content/analysis.tex
\section{Analysis}

\begin{figure}[!tbp]
\centering
\includegraphics[width=0.45\textwidth]{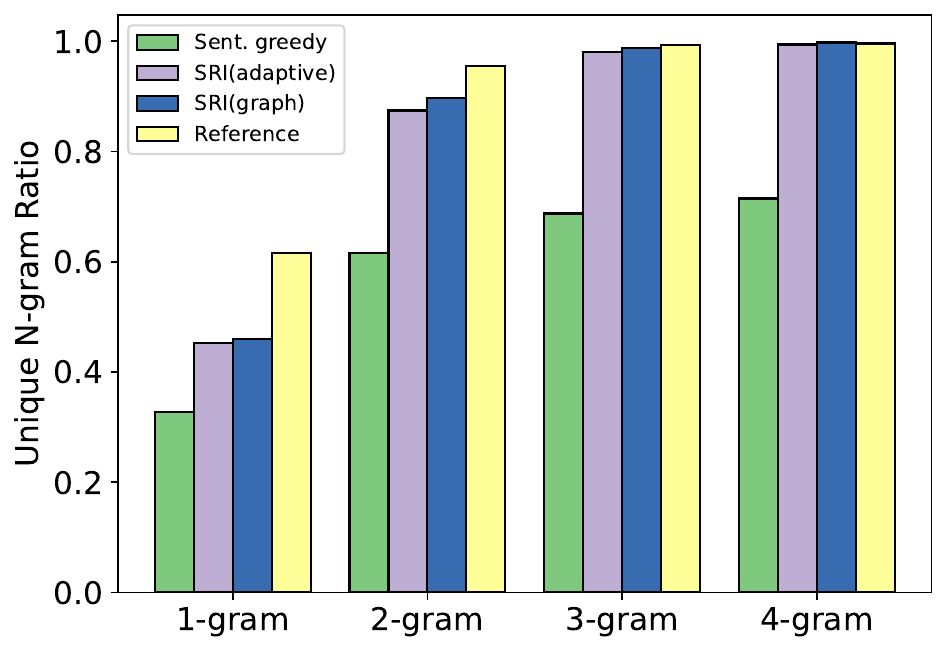}
\caption{Unique $n$-gram ratios ($n=1,2,3,4$) of the output summary by different methods on TAC-11.}
\label{fig:diversity}
\end{figure}

\subsection{Summary Diversity}
\label{sec:diversity}

Other than summary quality, we also test the effectiveness of our SRI in terms of the diversity of the output summaries. We present the unique $n$-gram ratios of output summaries under unsupervised and adaptive settings and the reference summary on the TAC-11 dataset in Figure~\ref{fig:diversity}. According to the results, our framework is extremely effective in reducing summary redundancy and increasing summary diversity under both unsupervised and adaptive settings. 

Compared to the ROUGE-F1 results, holistic inference with importance-diversity balancing measurement SRI increases both summary quality and diversity at the same time. The results suggest that considering summary diversity is beneficial in extractive summarization, especially in redundant cases like MDS and long document summarization. Our finding also verifies the crucial rule of effective modeling of sentence importance and similarity.

\subsection{Hyperparameter Sensitivity}

\begin{figure}[htbp]
\centering
\includegraphics[width=0.45\textwidth]{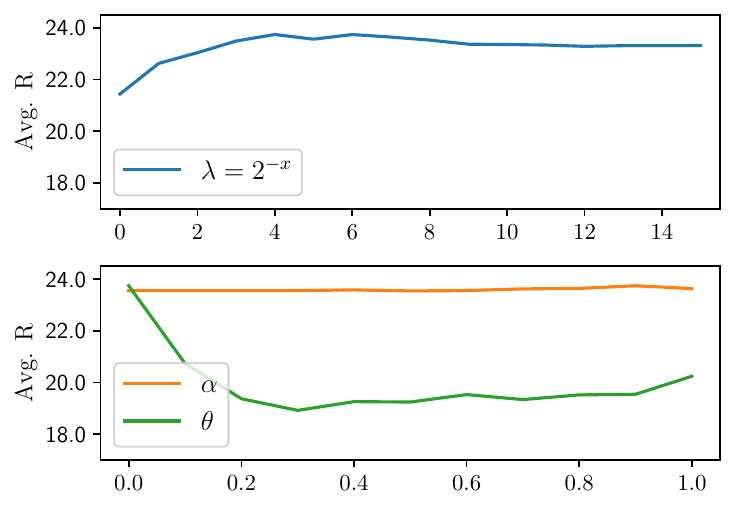}
\caption{Average ROUGE-F1 (w/o word limit) results with different hyperparameter values on TAC-11.}
\label{fig:hyper}
\end{figure}

To test the robustness of our proposed approaches, we study the hyperparameter sensitivity of our proposed methods.
The results are shown in \Cref{fig:hyper}.
The first plot shows the impact of balancing factor $\lambda$ in SRI.
The second plot shows the impact of $\alpha$, which balances the contextualized and TF-IDF sentence embedding and the edge weight threshold. The results show that our methods are relatively stable towards the hyperparameter values and could be easily adapted to unseen datasets.

\subsection{Inference Approaches Analysis}
\label{sec:inference_analysis}

\begin{figure}[htbp]
\centering
\includegraphics[width=0.42\textwidth]{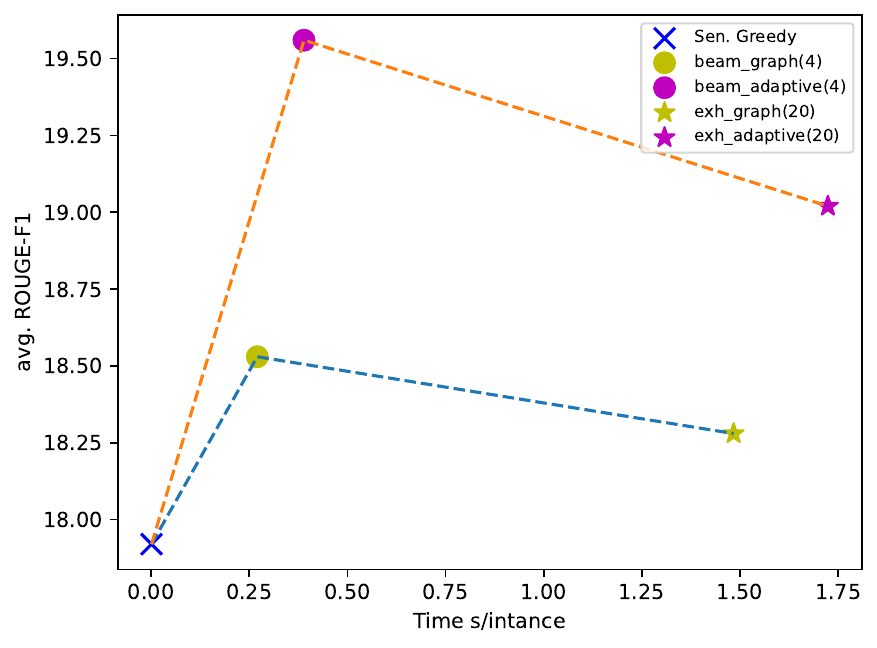}
\caption{Efficiency vs. average ROUGE (w/o word limit) scores  of different inference methods on TAC-11.}
\label{fig:decode}
\end{figure}

\input{tables/result_beam.tex}
We also compare the efficiency and effectiveness of different inference methods.
As in Figure.~\ref{fig:decode}, we compare sentence-level greedy search, set-level greedy search, set-level beam search (beam size = $4$), and set-level exhaustive search with pre-filtering as inference methods for both unsupervised and adaptive settings. We pick the filter size of $20$ here since the search space without filtering $C(N,K)$ is extremely large.
According to the results, all set-level inference methods outperform the sentence-level methods. This suggests that extracting summaries at a set level (holistic) is optimal over the common sentence-level setting that extracts sentences individually.
The finding is also consistent with the inherent performance gap between sentence-level and holistic extractors in \cite{zhong2020extractive}.

Moreover, we realize the set-level beam search and set-level exhaustive search achieve the comparable best performance.
However, set-level beam search speed-wise is much more efficient than set-level exhaustive search. We also show the effect of different beam sizes in Table~\ref{tab:beam}. The results indicate that a reasonably small beam size achieves the best ROUGE results, which are both effective and efficient. To conclude, set-level beam search with SRI shows the best overall performance.

%% file: tables/result_beam.tex
\begin{table}[t]
\setlength{\tabcolsep}{2.2pt}
\renewcommand{\arraystretch}{1.25}
\centering
\begin{small}
\begin{tabular}{l|ccccccc}
\hline
\textbf{Beam Size}  & \textbf{2} & \textbf{3} &  \textbf{4} & \textbf{5} & \textbf{6}& \textbf{7}  &\textbf{8} \\
\hline

ROUGE-1 & 33.43 & {33.65} & 33.62 & \textbf{33.76} & 34.72 & 33.64 & 33.67\\ 
ROUGE-2 & 7.71  & \textbf{8.00} & 7.87 & 7.93 & 7.84 & 7.86 & 7.85\\
ROUGE-L* & 28.74 & {29.03} & 28.99 & \textbf{29.10} & 29.01 & 28.94 & 29.01\\

\hline
\end{tabular}
\end{small}
\caption{
ROUGE-F1 (w/o word limit) results of SRI-beam with different beam sizes on TAC with $\lambda = 0.125$.
}
\label{tab:beam}
\end{table}

%% file: content/conclusion.tex
\section{Conclusion}

This paper proposes a holistic framework for unsupervised multi-document extractive summarization. Our framework incorporates the holistic beam search inference methods and SRI, a holistically balanced measurement between importance and diversity. We conduct extensive experiments on both small and large-scale MDS datasets under both unsupervised and adaptive settings and the proposed method outperforms strong baselines by a large margin. We also find that balancing summary set importance and diversity benefits both the quality and diversity of output summaries for MDS.